\documentclass[runningheads]{llncs}
\usepackage{graphicx}
\graphicspath{ {./img/} }
\usepackage{cite}

\begin{document}
\title{How to choose an Explainability Method? Towards a Methodical Implementation of XAI in Practice}
\titlerunning{Towards a Methodical Implementation of XAI in Practice}

\author{Tom Vermeire\inst{1}\and
Thibault Laugel\inst{2}\and
Xavier Renard\inst{2}\and
David Martens\inst{1} \and Marcin Detyniecki\inst{2,3,4}}
\authorrunning{Vermeire et al.}
%
\institute{University of Antwerp, Prinsstraat 13, 2000 Antwerp, Belgium \email{tom.vermeire@uantwerp.be} \and
AXA, Paris, France\\ \email{thibault.laugel@axa.com, xavier.renard@axa.com} \and
Sorbonne Université, CNRS, LIP6, F-75005, Paris, France \and
Polish Academy of Science, IBS PAN, Warsaw, Poland}
\maketitle              
\begin{abstract}
Explainability is becoming an important requirement for organizations that make use of automated decision-making due to regulatory initiatives and a shift in public awareness. Various and significantly different algorithmic methods to provide this explainability have been introduced in the field, but the existing literature in the machine learning community has paid little attention to the stakeholder whose needs are rather studied in the human-computer interface community. Therefore, organizations that want or need to provide this explainability are confronted with the selection of an appropriate method for their use case. In this paper, we argue there is a need for a methodology to bridge the gap between stakeholder needs and explanation methods. We present our ongoing work on creating this methodology to help data scientists in the process of providing explainability to stakeholders. In particular, our contributions include documents used to characterize XAI methods and user requirements (shown in Appendix), which our methodology builds upon.

\keywords{Explainable Artificial Intelligence  \and Interpretable Machine Learning \and Stakeholder needs \and Methodology.}
\end{abstract}
%
%
%


\section{Introduction}

Businesses are increasingly turning to machine learning systems to automate and enhance their operations and decision-making. By making use of complex modeling techniques, they are able to create models with high and sometimes super-human predictive performance. However, given their complexity, these models are often used as black-boxes for which it is unclear how predictions are made.

As a consequence, the business environment is and will increasingly be confronted with demands regarding explainability for various reasons. On the one hand, companies face internal challenges with the adoption of automated decision making systems due to limited explainability (trust in model, business insights, internal control)~\cite{arrieta2020, samek2019}. Moreover, the question of the impact of ML/AI models on business processes and the society is more salient than ever: external players will oblige companies to put explainability on the agenda as a priority. 

Hence, data scientists are and will be increasingly confronted with explainability demands from stakeholders beyond the traditional focus on predictive performance. This need of explainability ignited a whole new research field, often referred to as explainable artificial intelligence (XAI) or interpretable machine learning (IML). Over the past years, a multitude of methods has been proposed to explain machine learning models and decisions thereof~\cite{adadi2018, arrieta2020, molnar2020, 2018guidotti}. These methods significantly differ from each other in terms of output (explanation) and the way these explanations are generated and can therefore be more or less suitable for a specific use case and/or stakeholder.
There is thus a risk to fail to achieve a \textit{transparent} use of interpretability, which would be in contradiction with the initial objective of enlightening and empowering model's stakeholders. This issue has been noticed in previous works~\cite{bhatt2020, langer2021}, but a concrete methodology is lacking in the existing body of literature. The goal of this paper is to make an argument for the need of prescriptive guidance to match stakeholder needs with an appropriate explanation method. Moreover, we provide an overview of our ongoing work to introduce a methodology which is a first step towards bridging this gap.

This paper is divided in two parts.
The first one is devoted to the state of the art on implementing explainability in practice, from a user perspective and from a machine learning perspective.
We highlight the absence of thorough and sufficient cross-referencing between these two fields of research and in particular a clear methodology for a machine learning practitioner.
The second section proposes a reflection and a sketch of such a methodology for a sound scientific implementation of explainability in practice.


\section{Implementing explainability: current state of the literature}
\label{sec:current_state}

The question of practically using machine learning explainability has been poorly covered in the existing literature.
One notable exception is Bhatt et al.~\cite{bhatt2020}, which study the use of explanation methods in practice and thus show that, currently, they are mainly used by machine learning engineers in an \textit{ad hoc} way as sanity checks for the models they build and deploy.
One reason identified by the authors is that organizations lack frameworks for making decisions regarding explainability, leaving these methods only understandable by people with a background in machine learning and obscure to others.
This lack of focus on the perspective of the stakeholders to whom explainability is provided has also been observed by Langer et al.~\cite{langer2021}, who argue that more extensive stakeholder involvement is needed for the development and selection of explanation methods. 
They provide a conceptual model of the problem space in which they make a distinction between the explainability needs from stakeholders (so-called desiderata) and the explanation method that is used to address them. In their model, the success of an explanation method depends on the extent to which this method satisfies these needs and the careful consideration of these needs is therefore crucial. 

However, as the rather user/business-oriented explainability needs and the rather machine learning-oriented properties of explanation methods are completely different in nature, the mapping between the two is not straightforward, leaving the question of the use of XAI in practice unanswered.

Below, we give an overview of relevant work regarding the implementation of interpretability, which can thus be decomposed into two objectives: the collection of explainability needs from stakeholders and the selection of an appropriate explanation method. We conclude the section with a discussion on the lack of methodological guidance provided by the literature.

\subsection{Understanding stakeholder needs}
Typical categorizations of stakeholders are based on their role in an organization~\cite{tomsett2018, preece2018, bhatt2020, langer2021}, their machine learning experience~\cite{yu2018} or a combination of the two~\cite{suresh2021}. Also for the categorization of the stakeholder needs regarding explainability, different propositions are made in the literature. Some authors mention possible high-level goals of explainability, such as model debugging, monitoring etc.~\cite{bhatt2020} or revealing (un)known (un)knowns~\cite{preece2018}. Langer et al.~\cite{langer2021} provide a list of more detailed needs such as privacy, fairness, legal compliance, etc. Suresh et al.~\cite{suresh2021} propose an hierarchical framework of so-called tasks (e.g., understand influence of features on output), objectives (e.g., model debugging and compliance) and goals (e.g., trust and understanding) that contribute to each other~\cite{suresh2021}. 

In order to gather the needs from a stakeholder in a specific use case, different methods are introduced. Most of them take an approach stemming from information systems and software development research, where the collection of user needs is a well-known and well-studied problem. 
Köhl et al.~\cite{kohl2019} provide an analytical framework to elicitate and specify explainability needs and consider them as non-functional requirements which should be satisficed instead of satisfied. These can be translated into Softgoal Interdependency Graphs to represent their relationships. Other scholars claim that data scientists might have difficulties to directly discuss XAI solutions with stakeholders and, therefore, propose to rather focus on what a stakeholder wants to achieve with explainability~\cite{wolf2019, cirqueira2020}. In essence, they try to get a good understanding of the stakeholder's background, capabilities and goals. In this way, they create textual and/or visual scenarios that describe where and when explainability is needed. 
Another approach is taken by Liao et al.~\cite{liao2020}, who assume that an explanation can be seen as an answer to one or more questions of the stakeholder. They provide a set of possible questions for which the stakeholder might require answers in the form of an explanation. Eiband et al~\cite{eiband2018} propose a process consisting of different (iterative) phases which incorporates the position of different stakeholders and in which the first phases focus on what must be explained and the later phases look into how this can be explained. 

\subsection{Specifying explanation method properties}
On the other side of assessing stakeholder needs, different scholars propose a set of general characteristics or requirements to describe an explanation method. These allow to create a generalized identity card for explanation methods which can serve as documentation and a means to compare them. Both Hall et al.~\cite{hall2019} and Sokol and Flach.~\cite{sokol2020} introduced a framework of explanation method characteristics that can be filled in for a specific explanation method. 

The framework of Hall et al.~\cite{hall2019} consists of characteristics that are divided over the dimensions effectiveness, versatility, constraints, explanator types and categories, explanation properties and personal considerations. 
Sokol and Flach~\cite{sokol2020} provide a framework with five dimensions of so-called requirements of an explanation method. The functional requirements consists of requirements that determine whether it is practically feasible to use an explanation method for a specific use case. The operational requirements relates to the interaction of users with the system and their expectations. The usability requirements consists of properties that are of importance for the receiver of explanations. Finally, the safety and validation requirements respectively focus on aspects as privacy and security, and the validation of the explanation method. 

Without going into further detail about the individual explanation method properties in both frameworks, it can be argued that these dimensions not only cover generic properties of explanation methods but also (possible) stakeholder needs (e.g., constraints and personal considerations in Hall et al.~\cite{hall2019}, and operational and usability requirements in Sokol and Flach~\cite{sokol2020}). Compared to the stakeholder needs categorizations we mentioned above, the proposed templates do not give an exhaustive overview of all possible (future) needs, so certain of them might not be assessed and/or addressed. Moreover, describing how an explanation method addresses a specific need in a general way might limit practical usability of the framework. For instance, privacy is a broad concept that might imply different needs in different use cases. It can be about disclosure of the model, certain instances, certain features, etc. Similarly, the complexity of an explanation is a concept that can (simultaneously) cover different aspects. It can be impacted by the size of an explanation, its format, the type of features used in it, etc. Filling in this type of characteristics for a specific approach might lead to a description that is too general to confront with a specific use case. Another option is to fill in the template completely tailored to the use case at hand, which limits reusability and therefore to redo the exercise in each use case. 

\subsection{There is a lack of methodological guidance}
Given the explainability needs of a stakeholder and the properties of an explanation method, a logical next step is to assess whether the properties are capable of satisfying the needs. However, this is a cumbersome task, since there is no clear mapping between the properties and the needs. 
First, the explainability needs of a stakeholder and the explanation method properties are from a completely different nature. The former come from the domain of the (business) stakeholder and focus on information needs and domain constraints such as privacy and complexity. The latter rather relate to the XAI domain and focus more on the technical (algorithmic) details of an explanation method. As both sides use their respective jargon, a translation in between is considered necessary.   
Second, there might be a many-to-many relationship between needs and properties. A single property of an explanation method might contribute to multiple stakeholder needs, while the fulfillment of a single stakeholder need can be impacted by multiple and seemingly unrelated explanation method properties. 

To our knowledge, the XAI literature lacks prescriptive work on how to perform the mapping between explainability needs and explanation method properties. Approaches on gathering explainability needs typically ignore the properties of explanation methods or only discuss it to a limited extent (e.g., Liao et al.~\cite{liao2020} map their questions to a high-level taxonomy of explanation methods). More general frameworks to select explanation methods do mention the fact that a translation or mapping between needs and properties is needed, but provide no concrete guidance on how this can be done. For instance, Hall et al.~\cite{hall2019} only mention the mapping between explanation methods and stakeholder needs as a step in their methodology. Langer et al.~\cite{langer2021} provide some hypothetical scenarios to inspire future research on the mapping. It can be argued that creating a generalized identity card for an explanation method is a useful tool for data scientists to select an appropriate approach for a use case. However, we believe that this identity card should consist of characteristics that are closely related to the details of the explanation generation process, irrespective of potential needs that might be satisfied by them.   

In this work, we make a first step towards bridging this gap by proposing a methodology to translate explainability needs from stakeholders to the lower-level properties of explanation methods. This methodology aims at  guiding data scientists in the process of providing explainability to business stakeholders.


\section{Proposed methodology}
\begin{figure}[h]
\centering
\includegraphics[width=\linewidth]{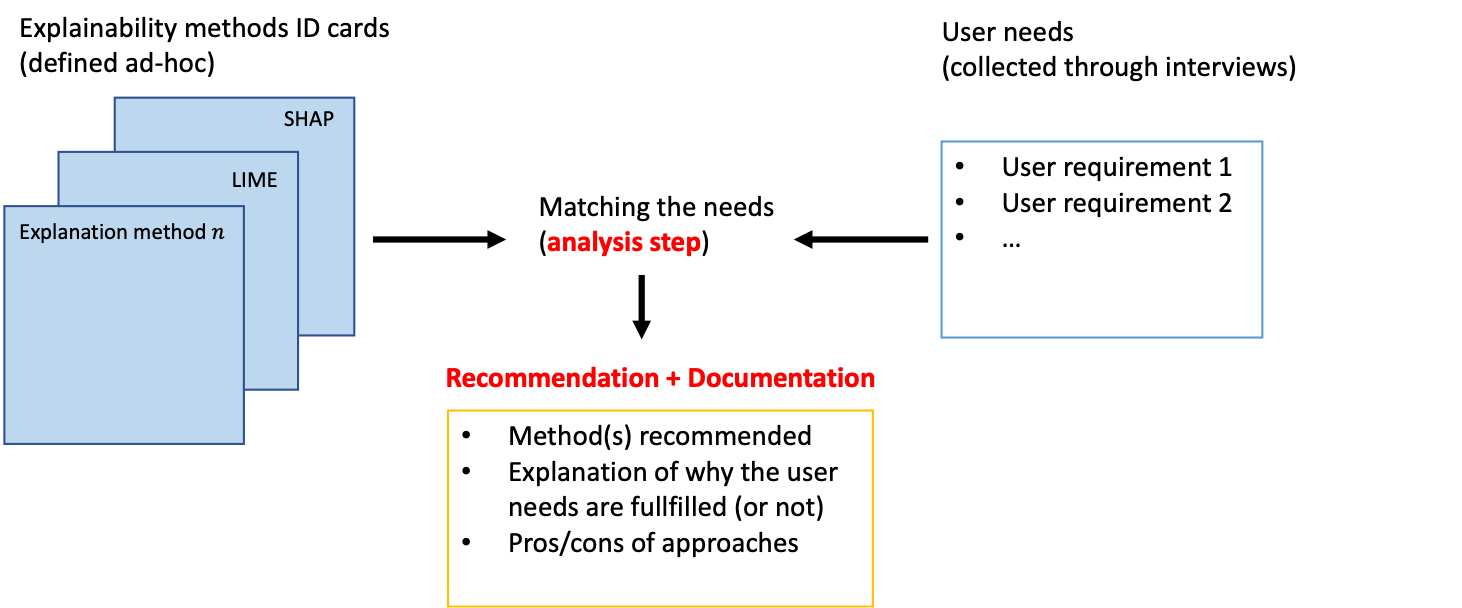}
\caption{Illustration of the proposed methodology and its different components.}
\label{fig:methodology}
\end{figure}

The goal of this work is to provide a (first step towards a) methodology that can guide a data scientist in tackling an explainability use case. We revisit both the  explainability needs and explanation method properties, in order to get them at a level of detail that allows to match them. Based on existing work, we define a template for them that can be used to document both aspects. The current versions of the templates are included in the appendix of this paper. Moreover, we plan to provide guidance on how the templates can be completed and how they can be used to find an appropriate explanation method for an explainability use case. An overview is shown in Figure~\ref{fig:methodology}.

\subsection{Explanation method properties}
Using Hall et al.~\cite{hall2019} and Sokol et al.~\cite{sokol2020} as a starting point, we consider it valuable to create an identity card (later referred to as "ID card") to describe an individual explanation method in detail. However, we make a clear distinction with (and therefore not include) the explainability needs that might be satisfied by a method, thus differentiating it from currently existing versions~\cite{hall2019,sokol2020}. The current version of our identity card, which is included in Appendix~\ref{app:properties} for the sake of clarity, consists of properties that are organized into the four dimensions discussed below. 

\subsubsection{Compatibility properties}
This set of properties allows to determine whether it is practically feasible to apply an explanation method to a certain use case. It covers aspects such as the type of the task, the scope of an explanation, the model and the training data. 

\subsubsection{Explanation properties}
This set of properties describes the explanation that is produced by an explanation method. This comprises the information that is provided, the language and the relationship with the model and the real world.

\subsubsection{Method usage}
This section describes how an explanation method can be used in practical terms. It discusses whether code is available and how it can be used.

\subsubsection{Process properties}
This set of properties describe the process an explanation method uses to produce explanations. It covers aspects that are not necessarily visible from an explanation itself, but that can play an important role in which final explanation is obtained. 


\subsection{Explainability needs}
We also propose a card that gives a detailed overview of stakeholder needs. To construct this document, we started from the existing literature discussed in Section~\ref{sec:current_state}. Subsequently, we selected needs that were considered relevant and detailed enough to work with throughout a brainstorm between the authors of this paper. For these needs, we created a questionnaire that can be used to reveal the needs from a stakeholder. The resulting needs card and questionnaire were further refined by applying them to actual use cases and by integrating the feedback. The current version of the needs card, which is included in Appendix~\ref{app:needs}, is organized in the four dimensions discussed below. The corresponding questionnaire is outlined in Appendix~\ref{app:questions}.

\subsubsection{Use case context}
This dimension gives an overview of the use case for which explainability is required. It contains the business process in which the ML system fits and a detailed description of the ML system itself.

\subsubsection{Stakeholder}
In order to provide explainability to a stakeholder, it is necessary to have a good understanding of his role and his background, both regarding the business domain and machine learning in general. 

\subsubsection{Stakeholder needs}
This dimension bundles the aspects that express what information a stakeholder wants to derive from the explanation. In traditional information systems terms, this dimension considers the functional requirements of the explanation.

\subsubsection{Stakeholder constraints}
This dimension bundles aspects that put restrictions on the explanation (process) that are relevant to the stakeholder. In traditional information systems term, these can be called the non-functional requirements of an explanation.


\subsection{Information collection}
The completion of the template for the explanation method properties is considered a task the research community that is in charge of providing explanation methods. Existing and future explanation methods should be subjected to a thorough investigation in order to assess their properties. To this end, experts can rely on existing documentation (research papers, reports, code base etc.) and experience by experimenting with the method. We acknowledge this is a cumbersome task, but we believe that this documentation, if properly composed, will form a valuable and comparable knowledge base that can be reused for future use cases.

To gather the information on stakeholder needs for a specific use case, the XAI expert should probe the relevant stakeholder. Since certain aspects might be more technical in nature, it is considered necessary to also involve the people that are responsible for the development and deployment of the model that is used in the use case (if they are not the stakeholder themselves). Our questionnaire can be used as a tool to guide this process, but also more approaches that are more tailored to the specific stakeholder might be explored (see the literature we discussed in Section~\ref{sec:current_state}).

\subsection{Matching stakeholder needs with explanation method properties}

After the collection and documentation of the stakeholder needs, the next step is to confront them with the knowledge base of detailed explanation method properties in order to select the (most) appropriate explanation method(s). As argued in Section~\ref{sec:current_state}, their is no clear relation between stakeholder needs and corresponding explanation method properties that can be generalized over use cases. However, the separation of the two sides makes their different nature explicit and aids to prevent the oversimplification of the problem. This should encourage the XAI expert to perform a thorough analysis and resulting mapping for each use case. While acknowledging that the current specifications of the stakeholder needs and explanation method properties are not final and should be further revised and refined, we believe they provide the XAI expert with useful tools in that process.

We advocate to document in detail the analysis and the resulting choice of explanation method(s). This encourages the XAI expert to show to what extent the stakeholder needs are addressed and to point at disadvantages or risks that are related to using an specific explanation method. This should result in a substantiated and nuanced advice for a specific use case. 


\section{Ongoing and Future Research}

In this paper, we discussed our on-going work of designing a methodology to help XAI stakeholders choosing an explainability method. Besides identifying and discussing the issue and the current state of the art, our contributions include a prototype of ID cards and questionnaire to identify user needs (both provided in Appendix).

The following step in this research project is to apply this methodology in practice. 
First, we plan on enriching the current knowledge base of ID cards with the most common explanation methods and their properties.
Alongside, we are currently conducting interviews with stakeholders working on specific real-life use cases requiring interpretability. These semi-structured interviews allow us to validate and refine the current stakeholder needs specification and the corresponding questionnaire, which is still an ongoing process. By considering use cases from diverse domains and involving different (types of) stakeholders (e.g. data scientists, non-expert customers, auditors...), we plan on covering as many different needs as possible. In particular, we also pay close attention to the extent to which stakeholders generally can or cannot answer questions related to a specific need. 

The next step is to perform the analysis to match the stakeholder needs with an appropriate explanation method. We will do so for each use case collected in the interviews and present a (set of) explanation method(s) that can be considered a suitable option. Subsequently, the proposed solution(s) will be discussed with and demonstrated to the stakeholder as a validation step. This will allow us to assess the effectiveness and discover the limitations of our methodology. 
By submitting this paper to the XKDD workshop, we hope to have interesting discussions with leading XAI researchers to challenge our in-the-making  methodology and thus help solve what we believe to be a fundamental issue of the field today.


\bibliographystyle{splncs04}
\bibliography{references}


\appendix 
\section{Explanation method properties (ID cards)}
\label{app:properties}

\subsection{Compatibility properties}

\begin{itemize}
    \item Problem supervision level: unsupervised, supervised, reinforcement
    \item Problem type: classification, regression, clustering, etc.
    \item Explanation target: data, model, predictions
    \item Explanation scope: instance-level or global
    \item Applicable model class: model-agnostic or model-specific
    \item Ante-hoc (by design) of post-hoc (a posteriori)
    \item Compatible data types: tabular, image, text, etc.
    \item Compatible feature types: numerical, ordinal, categorical
    \item Training data-agnostic or training data access needed
\end{itemize}

\subsection{Explanation properties}
\begin{itemize}
    \item Explanation family: association between antecedent and consequent, contrasts and differences, causal mechanisms
    \item Explanatory medium: representation of explanation
    \item Explanation domain: original feature domain or transformed domain
\end{itemize}

\subsection{Method usage}
\begin{itemize}
    \item Code availability: source code
    \item Inputs: detailed description of arguments
    \item Outputs: detailed description of return values
\end{itemize}

\subsection{Process properties}
\begin{itemize}
    \item Computational complexity: big-O notation or based on empirical evaluation
    \item Fidelity: extent to which underlying model is closely mimicked 
    \item Optimality: exact or heuristic explanation generation
    \item Sparsity: size of explanation
    \item Coverage: guarantee of explanation
    \item Plausibility: extent to which explanation is within data manifold
    \item Consistency: extent to which similar instances get similar explanations
    \item Stability: extent to which explanation generation is subject to randomness
    \item Interactiveness: extent to which end user can interact with explanation generation
\end{itemize}


\section{Explainability needs}
\label{app:needs}

\subsection{Use case context}
\begin{itemize}
    \item Business process and role of ML system
    \item Model inputs (data and features)
    \item Model output
    \item Model type
\end{itemize}

\subsection{Stakeholder}
\begin{itemize}
    \item Role in the organization and the business process
    \item Background regarding machine learning in general and regarding the application domain
\end{itemize}

\subsection{Stakeholder needs}

\begin{itemize}
    \item Information needs: information that must be provided by explanation to answer questions from stakeholder
    \item Language of the explanation
    \begin{itemize}
        \item Format type: textual/visual/…
        \item Feature specificity: fine-grained features vs. higher-level features
        \item All model features vs. subset of features
    \end{itemize}   
    \item Truthfulness to model: extent to which an approximation of the model is allowed.
    \item Faithfulness to real world
    \begin{itemize}
        \item If actionability: feasibility of actions
        \item If case: realisticness of case (in data distribution)
    \end{itemize}

\end{itemize}

\subsection{Shared stakeholder constraints}
\begin{itemize}
    \item Timeliness: time frame in which explanation must be provided
    \item Explanation complexity: degree of difficulty that can be processed by stakeholder within given time frame
    \item Privacy / Intellectual property: disclosure of instances, features or model reasoning
    \item Robustness: extent to which explanations can differ between instances, models and over time
\end{itemize}

\clearpage
\section{Questionnaire to reveal needs}
\label{app:questions}

\subsection{Use case context}

\begin{itemize}
    \item What is the business process and where does the ML system fit?
    \item What are the model inputs?
    \item What are the model outputs?
    \item What type of model is used?
\end{itemize}

\subsection{Stakeholder}

\begin{itemize}
    \item Can you describe the stakeholder and his role in the business process?
    \item Background:
    \begin{itemize}
        \item Do the individual model features make sense to the stakeholder?
        \item What is the ML background of the stakeholder?
    \end{itemize}
\end{itemize}

\subsection{Stakeholder needs}

\begin{itemize}
    \item Information needs 
    \begin{itemize}
        \item Does stakeholder want general insight in the system or insight for a specific case?
        \item What does stakeholder want to achieve/do with the explanation?
        \item If global:
        \begin{itemize}
            \item Does stakeholder want to know which features are most/least important?
            \item Does stakeholder want to have insight in the general logic of the system?
            \item Does stakeholder want to have (a-)typical instances?
        \end{itemize}
        \item If local:
        \begin{itemize}
            \item Does stakeholder want to know the logic of the model regarding this decision?
            \item Does stakeholder want to know which features contributed the most/least?
            \item Does stakeholder want information on how to change the decision (actionability)?
            \item Does stakeholder want to know how changes to the instance affect the decision?
            \item Does stakeholder want to see comparable or opposite cases?
        \end{itemize}
    \end{itemize}
    
    \item Language of the explanation
    \begin{itemize}
        \item Does stakeholder prefer a textual explanation, visual explanation, rule etc.? 
        \item Does stakeholder want an explanation in terms of the individual features, or is an higher abstraction level considered sufficient or desirable?
        \item Is it allowed to only consider a subset of understandable features and ignore other features that might also have impacted decision-making?
    \end{itemize}
    
    \item Truthfulness to model
    \begin{itemize}
        \item Is using an inherently transparent model for this use case an option?
        \item Is it allowed to make an approximation, meaning that the explanation might not be completely truthful to the model?
    \end{itemize}
    
    \item Faithfulness to real world
    \begin{itemize}
        \item In case stakeholder wants actionability: Do the guidelines need to be actionable/realistic for the stakeholder?
        \item In case stakeholder wants case or counterfactual: Should instance be close to data distribution
    \end{itemize}
\end{itemize}

\subsection{Stakeholder constraints}
\begin{itemize}
    \item Timeliness
    \begin{itemize}
        \item Does stakeholder always need an explanation?
        \item If not, how quickly does stakeholder want an explanation after requesting one?
    \end{itemize}
    
    \item Explanation complexity
    \begin{itemize}
        \item How large can an explanation be?
        \item How much time does stakeholder have to process the explanation?
    \end{itemize}
    
    \item Privacy / IP
    \begin{itemize}
        \item Are there model features that cannot be shown?
        \item Can (a part of) the model logic be disclosed?
        \item Can other data instances be shown?
    \end{itemize}
    
    \item Robustness
    \begin{itemize}
        \item Should similar instances have similar explanations?
        \item Should explanation for certain instance or model be the same if model changes over time?
    \end{itemize}
\end{itemize}

\end{document}